\documentclass[conference]{IEEEtran}
\IEEEoverridecommandlockouts
\usepackage{cite}
\usepackage[utf8]{inputenc}
\usepackage{graphicx}
\usepackage{amsmath}
\usepackage[version=4]{mhchem}
\usepackage{siunitx}
\usepackage{longtable,tabularx}
\usepackage{subcaption}
\usepackage{epstopdf}
\usepackage{algorithm}%
\usepackage{algorithmicx}%
\usepackage{algpseudocode}%
\usepackage{tabularx}
\usepackage{booktabs} % \**rule table line package
\usepackage{multirow}
\usepackage{arydshln} % dashline package
\usepackage{url}
\usepackage{float}
\usepackage{acronym}
\usepackage{fancyhdr}
\fancypagestyle{specialfooter}{%
  \fancyhf{}
  
  \fancyfoot[L]{\footnotesize This work was presented at the AIAA Aviation 2023 Forum. \\ Copyright @ Souma Chowdhury}
  \fancyfoot[C]{1}
}
\usepackage{hyperref} % makes references hyperlinks
\usepackage{textcomp} % Symbol Expansion Package
\def\BibTeX{{\rm B\kern-.05em{\sc i\kern-.025em b}\kern-.08em
    T\kern-.1667em\lower.7ex\hbox{E}\kern-.125emX}}

\renewcommand{\footnoterule}{%
  \kern -3pt
  \hrule width \columnwidth height 1pt
  \kern 2pt
}

\begin{document}

\title{Faster Thermal Profiling of a Lunar Rover with \\Machine Learning Adapted Finite Difference Model}

\author{\IEEEauthorblockN{Samuel Weber$^*$ \thanks{$^*$ Ph.D. Student, Department of Mechanical and Aerospace Engineering, AIAA Student Member}}
\IEEEauthorblockA{\textit{University at Buffalo}\\
Buffalo, New York, 14260 }
\and
\IEEEauthorblockN{Zaki Hasnain$^\dag$ \thanks{$^\dag$ AIAA Member}} 
\IEEEauthorblockA{\textit{Zij Sciences} \\ Ashburn, VA, 20147 }
\and
\IEEEauthorblockN{Souma Chowdhury$^\ddag$ \thanks{$^\ddag$ Professor, Mechanical and Aerospace Engineering, AIAA Associate Fellow, Corr. author, soumacho@buffalo.edu} 
\thanks{This work is accepted to be presented at the AIAA AVIATION 2026 Forum.}
% \thanks{Copyright \copyright 2024 ASME. Personal use of this material is permitted. Permission from ASME must be obtained for all other uses, in any current or future media, including reprinting/republishing this material for advertising or promotional purposes, creating new collective works, for resale or redistribution to servers or lists, or reuse of any copyrighted component of this work in other works}
}
\IEEEauthorblockA{\textit{University at Buffalo}\\
Buffalo, New York, 14260 \\
}
}

\thispagestyle{specialfooter}
% \author{Zaki Hasnain\footnote{AIAA Member}}
% \affil{Zij Sciences, Ashburn, VA, 20147}
% \author{Souma Chowdhury\footnote{Associate Professor, Department of Mechanical and Aerospace Engineering, Co-Director, Center for Embodied Autonomy and Robotics, AIAA Associate Fellow, Corresponding author. Email: soumacho@buffalo.edu}}
% \affil{University at Buffalo, Buffalo, NY, 14260}

\maketitle
\begin{abstract}
Autonomous space systems operating in extreme thermal environments require accurate and efficient thermal modeling to support both pre-mission systems design and during-mission autonomy. In the context of lunar rovers, large temperature gradients, radiative heat transfer, and variable surface conditions make reliable prediction of their thermal profile particularly challenging. High-fidelity physics-based simulations provide accurate thermal predictions but are computationally expensive, while simplified models and precomputed lookup approaches often lack sufficient accuracy. Physics-informed machine learning (PIML) offers a promising alternative by combining data-driven models with embedded physics knowledge or models.
This paper presents a PIML framework for thermal analysis of a simplified lunar rover with internal heat sources, where ML enables environment-adaptive coarse meshing. More specifically, the proposed architecture integrates a transfer neural network (TNN) that adaptively determines 3D finite difference (FD) nodalization based on thermal loads and initial conditions, making more accurate coarse mesh calculations possible. A differentiable FD thermal simulator is embedded within the framework to enforce physical consistency and enable efficient training, while an upscaling layer is used to recover high-resolution temperature fields from the coarse-grid solution.
The performance of the proposed PIML approach is evaluated against high-fidelity (fine mesh) and low-fidelity (fixed coarse mesh) physics-based models, as well as a purely data-driven artificial neural network (ANN). Results show that the PIML framework improves prediction accuracy by 50\% and 39\% relative to lower-fidelity (coarse mesh) physics and ANN models, respectively, while maintaining physically consistent thermal distributions. Computationally, it is also 3$\times$ faster compared to the high-fidelity simulations, thereby balancing accuracy and efficiency in thermal modeling of rover systems.
\end{abstract}


\renewcommand{\thesection}{\Roman{section}}
\section{Introduction} \label{S:intro}
Planetary rovers play a vital role for NASA \cite{keane2024endurance, baker2025endurance, smith2022viper}, other space agencies \cite{quantin2021oxia, casanova2025tenacious}, and the private sector \cite{coggins2024artemis, witze2023moon, casanova2025tenacious}, especially with increased focus on autonomous and potential crewed exploration of the lunar and Martian surfaces. In addition to the mobility, power, thermal, and telecommunication subsystems necessary for navigation and safe operation, these rovers carry complex science instrumentation payloads. For example, NASA's Curiosity rover contained cameras, spectrometers, radiation detectors, an environment monitoring station, and soil sample acquisition systems \cite{quattrocchi2022curiosity}. These engineering and science systems must withstand harsh and dynamic thermal environments to accomplish complex mission tasks. For example, surface conditions include extreme variations in thermal loads due to solar incidence angle, shadowing effects from surface features, dust, day-night transitions, and radiation from nearby terrain such as large rocks. Therefore, firstly, engineers must design thermally robust rovers within a mission's power, mass, and concept of operations (ConOps) constraints. In particular, on the lunar surface, the lack of a substantial atmosphere leaves the surface exposed to space, causing it to be treated as a near-vacuum environment, which equates to radiation being the primary heat transfer mechanism \cite{haviland2021lunar}. As such, environmental conditions and the ConOps will dictate whether a rover requires cooling or heating to maintain required operating temperatures for a given component. Thus, these rovers require a sufficient passive or active thermal control system (TCS) to account for this variability in thermal loads. To design this TCS, the thermal engineers must have the ability to perform a complete thermal profiling of the rover. Now, secondly, knowledge of thermal profiles or the ability to predict them on the go are also required during spacecraft operations, either to support onboard autonomy or off-board planning  -- e.g., task and path planning, or active heating/cooling decisions. This paper presents the development and initial findings on a new physics-informed machine learning approach to rover thermal profiling that seeks to provide the necessary balance between accuracy/robustness and compute cost to address the above stated design and operations needs.

Traditionally, there are two main classifications of approaches to model rover thermal profiles: simplified physics models or high-fidelity computational models. For the first class, these models often come with the main drawback of needing significant knowledge of the topic to make heuristic choices of parameters to simplify. These simplifications often are based on analytical solutions that may be found, but these often only apply to very idealized cases, such as transient, lumped, and steady-state one-dimensional processes that result in an ordinary differential equation \cite{jaluria2002heat}. Additionally, these simplifications can lead to larger confidence intervals and, in the rover context, result in conservative designs with suboptimal performance. Addressing the shortcomings of this approach is the other class, high-fidelity computational models that numerically solve the governing differential equations using, for example, finite difference or finite element schemes. As a consequence, such simulation approaches involve increased computational costs  \cite{jaluria2002heat}. While such costs are acceptable for ad hoc analysis, they are ill-suited for iterative or optimization-driven design cycles (likely causing mission development delays). Moreover, they are impractical for any sort of operations-level decisions that must adhere to strict decision-time horizons, and are not amenable to onboard deployment over memory-/processing--scarce systems in rovers. 

On the other end of the spectrum of models is the use of data-driven surrogate models. This approach aims to address high computational costs and serve as a useful tool for optimization and prediction \cite{ewim2021ann}. In addition, these models do not require extensive domain knowledge, as they are trained solely on given data \cite{mayerhofer2024comparison}. One disadvantage of this approach is its black-box nature. Necessary verification steps are challenging to perform as the weights and connections made by the model have little to no association with the underlying physics. Moreover, they are often brittle when generalizing over large input spaces. Because of these drawbacks, an emerging paradigm combines such data-driven approaches with physics models to constrain and/or enhance the predictions made. This class is broadly referred to as physics-informed machine learning (PIML), whose background and applications are discussed in the following subsection.

\subsection{Physics Informed Machine Learning (PIML) Methods}
There has been continued growth in interest in modeling methods that combine physics with data-driven Machine Learning (ML) models. These models have advantages such as being efficient with ill-posed and inverse problems, and have better generalization, extrapolation, and predictive performance on small dataset problems \cite{karniadakis2021piml}. By comparison, another common method is the use of artificial neural networks (ANNs), which have the benefit of adaptive learning, self-organization, and real-time computations due to their parallel computing ability \cite{qamar2023ann}. However, as these ANN models are solely data-based, the quality and amount of data greatly impact the final predictions made \cite{dayhoff2001ann}. To reduce the data requirements, physics-informed neural networks (PINNs) utilize known physics of the problem, with classical PINNs implementing physics in the loss function through the use of the physics partial differential equation as an additional loss \cite{mao2020pinn, raissi2019pinn, zhang2022pinn}. As they are similar to ANNs, they will have similar computational costs, but the additional physics loss function can enhance performance in some cases \cite{cai2021review}. Even with the addition of the physics loss function compared to ANNs, the trained weights have little physical meaning when interpreted. In addition, PINNs also give rise to significant training challenges \cite{haitsiukevich2023improved}. 

Another physics-based ML architecture is a hybrid PIML, that contain a compute-efficient (albeit low-/moderate-fidelity or partial) physics model within, rather than only in the loss function. This physics-based model offers potentially better generalization and helps mitigate the degradation that data-driven models have when conditions vary from those of the training data \cite{ewim2021ann}. In the past years, there have been many PIML architectures described in literature \cite{
kapusuzoglu2020hybrid,
Rufa2020HybridChem,
Matei2021Quadcopter,
Zhang2021MidPhyNet,
Maier2022MedicalImagingReview,
AnkobeaAnsah2022DieselEngine,
Rajagopal2022PETMRI,
Freeman2022MarineTurbine,
Lai2022ReflowOven,
Machalek2022EnergyModeling,
Choi2021FlightPrediction,
Chen2021HybridSpectroscopy,
Rai2021HybridDamage,
Den2024Battery,
sinha2025piml,
Zhaoqin2025,
de2024digital,
Iqbal2024, Behjat2020, Oddiraju2025}, which describes applications in modeling dynamic systems, robotic systems, material behavior, and battery health prediction, to name a few. Therein, the ML component often plays the role of adapting, correcting, and/or substituting aspects of the low/moderate-fidelity physics component, usually conditioned on the inputs. 

The PIML architecture described in \cite{oddiraju2024piml} is the predecessor to the one described in this paper, where the role of the ML component was to decide the mesh or grid coarseness settings for various parts of the physical system, subject to the environmental conditions (thermal loads). It combines an ANN with a partial physics model, with the main implementation being a data-driven model to map thermal inputs to those of the partial physics model, and then aligning its temperature outputs with those of a high-fidelity model. This provided more transparency than purely data-driven models. Scalability and efficiency of training were achieved by choosing Google JAX \cite{jax2018github} for the partial physics model, which is later integrated into the transfer networks implemented in PyTorch \cite{paszke2019pytorch}, thereby enabling end-to-end auto-differentiation of the PIML architecture.

\subsection{Research Objectives}

The primary goal of this paper is to develop a PIML architecture that predicts optimal finite difference (FD) nodalization based on internal and external thermal loads. This work builds upon \cite{oddiraju2024piml} by adding capabilities such as varying node (spatial) density and handling of internal loads, and thereof application to thermal analysis of a lunar rover. 
In practice, large-scale extraplanetary rover development involves multiple subsystem teams operating concurrently. For instance, design changes in one subsystem propagate across others, requiring iterative updates to inputs such as material properties and operating temperatures. Thermal analysis performed by the thermal control system (TCS) group further influences these design decisions, creating a feedback loop that can become a bottleneck due to the high computational cost of conventional tools such as Thermal Desktop and SINDA/FLUINT \cite{peabody2022thermal}.
This work aims to alleviate these challenges by providing an efficient, verifiable, and repeatable thermal analysis framework to support early-stage design and time-sensitive decision-making for onboard autonomy. The primary contributions of this work are as follows:
\begin{enumerate}
\item Development of a PIML architecture\textit{ where a neural network component predicts optimal nodalization (grid density) based on internal and external thermal loads (inputs) to be used by an explicit 3D FD model} in a manner that allows the average grid density to be several folds lower than that of a suited high-fidelity FD model -- thereby paving the way for \textit{increased compute efficiency}.
\item Implementation of an \textit{auto-differentiable finite difference thermal model} incorporating conduction and radiation, as well as \textit{load-downsampling and temperature-upscaling schemes} within the PIML architecture -- thereof enabling \textit{efficient training} and out-of-the-box deployment once trained. 
\item Demonstration of \textit{improved accuracy–efficiency trade-offs} through comparison with high-fidelity physics-based (FD), low-fidelity physics-based (FD), and purely data-driven (ANN) models across varying thermal conditions.
\end{enumerate}

The remainder of this paper is laid out as follows: In section \ref{S:method}, we describe the thermal simulator and the PIML architecture in detail. In section \ref{S:casestudy} we describe the lunar rover modeling case study, including the finite difference scheme used. Section \ref{S:results} describes our results, before moving on to section \ref{S:conclusion} containing our concluding remarks.

\section{PIML Thermal Modeling Framework} \label{S:method}
The PIML architecture relies on a differentiable thermal simulator to provide physics-based information for evaluating training losses. In addition, the framework compensates for the use of a coarse finite difference (FD) mesh by incorporating high-fidelity data during training. This section first outlines the structure of the thermal simulator, followed by a detailed description of the PIML model.

\subsection{Differentiable Thermal Simulator}
In this paper, we restrict the analysis to a finite difference (FD) thermal model, utilizing an explicit method for calculating heat transfer. Like \cite{oddiraju2024piml}, we opted to create this model using JAX, and software such as Thermal Desktop is not auto-differentiable and cannot be implemented into the PIML architecture, thereby enabling back-propagation. The rover model itself will be modeled as a simplified rover with the body being a lumped mass at an initial uniform temperature. External and internal heat fluxes will be applied, where the internal represents heat dissipated from equipment at certain locations or internal heat sources. The temperature will be calculated throughout the entire body. The environmental data are modeled based on data gathered from NASA's Apollo missions and other lunar data found in \cite{hamill2021lunar}.  The remainder of this subsection outlines the procedure for the simplified physics model.

All external and internal loads ($q$) or heat flux ($q''$) are considered to be constant over the duration of the simulation. This assumption is based on the relatively slow movement of the rover compared to the simulation time. Due to the choice to use an FD mesh, all external and internal loads and fluxes must first be discretized. Once done, loads are down-sampled to work with the sparse nodalization, which is a required input for the thermal simulation. A down-sampling scheme proposed in \cite{oddiraju2024piml} preserves the total energy input, and a similar scheme is used here and described in more detail in the following section. An explicit finite difference scheme is executed, producing an updated temperature profile. The transient heat transfer is modeled in 3D, while cylindrical rods are reduced to 1D representations.

The reference data used in the loss function are based on dense nodalization; thus, the predicted temperatures must be interpolated to match the data during training and inference. In this paper, we suggest adding an upscaling layer to interpolate and adjust temperatures, compared to classical methods, such as bilinear or bicubic interpolation.

\subsection{PIML Architecture}

\begin{figure*}[ht!]
    \centering
    \includegraphics[width=0.78\linewidth]{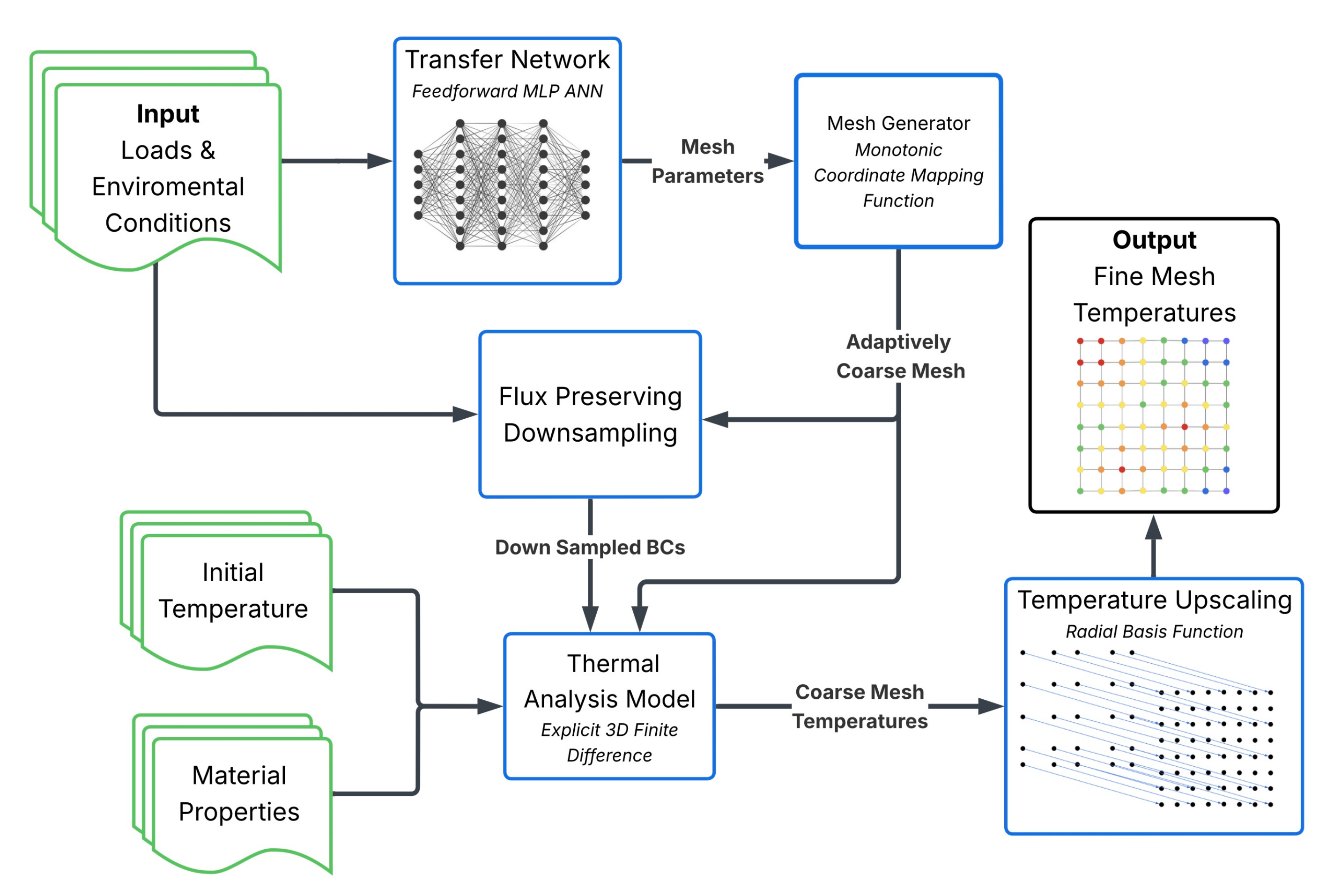}
    \caption{Model architecture of the proposed PIML model trained on Python simulator data}
    \label{fig:PIML}
\end{figure*}

Our PIML strategy employs sampling from high-fidelity simulations to improve the geometric nodal distribution. The input data are passed through a transfer neural network (TNN), outputting variables that can be used to generate the FD grid of node positions. Subsequently, the node positions are fed into the physics simulator along with material properties and loads/fluxes downsampled (to the coarse mesh informed by the transfer network). The FD model then outputs the temperature profile on the sparse grid mesh, which is then fed into an upscaling layer to provide the predicted temperatures on a fine grid mesh. Hence, the end outcome is a temperature profile at the same level of resolution as the high-fidelity grid, thereby allowing ease of use and training of the PIML model with a loss function that can directly compute the error without any further shifts or scaling. Figure \ref{fig:PIML} illustrates the framework just described. The rest of this section describes the main components of this framework, namely the transfer network, downsampling process, and upscaling layer, with the FD model described later in Section \ref{S:casestudy}. 

\subsubsection{Transfer Network}
For the transfer neural network (TNN), a feedforward Multilayer Perceptron (MLP) will be used. This method has been adopted in a wide range of applications, including classification, regression, and other machine learning tasks \cite{Dalatu2016ACS}, and was therefore chosen for implementation here. The inputs to this neural network will be the loads and environmental parameters, and it will predict the number of nodes per component and the spacing required on the rover. The overall purpose is to generate the nodalization of the finite difference grid that ultimately will be passed as an input to the embedded physics model. Equation \ref{eq: TNN} represents the formulation of the TNN that will be used in this paper. An example of the nodalization output generated by the TNN is provided in Appendix A to illustrate the spatial discretization predicted by the model.

\begin{equation}\label{eq: TNN}
\begin{split}
    &Y = f_{TNN}(X,\Theta) =   \\ & W^{(L)}\phi(W^{(L-1)}\phi(...\phi(W^{(1)}X + b^{(1)})... + b^{(L-1)}) +b^{(L)}
\end{split}
\end{equation}

\noindent where:
\begin{description}
\item  $Y$ is the finite difference grid
\item $X$ is the input thermal loads
\item $\Theta$ represents all learnable parameters
\item $\phi$ is the activation function 
\item $W$ is the weight of each layer
\item $b$ is the bias vector
\end{description}

\subsubsection{Down-sampling Process}
The thermal loads inputs are based on a high-fidelity dense node mesh; however, the physics model operates on a lower-fidelity (sparse) nodalization mesh. Mapping between these two requires special attention to the total amount of energy input into the system, as this would impact the outcome of the system. As such, traditional downsampling methods would not work. As first seen in \cite{oddiraju2024piml}, a downsampling scheme is proposed to preserve the total flux input into any surface. The algorithm detailed below outlines the procedure for a 2D face; however, the same process will apply to both 1D and 3D cases.

\begin{algorithm} 
\caption{Flux-Preserving Load Downsampling (Rectangular Faces)}
\begin{algorithmic}[1]
\State \textbf{Input:} Dense mesh $D$, Sparse mesh $S$, Loads on Dense mesh $Q_D$
\State \textbf{Output:} Interpolated loads on sparse mesh $Q_S$
\State
\Procedure{Downsampling}{$D$, $S$, $Q_D$}
    \State \textbf{Step 1:} Compute the corners of all elements in the reference dense mesh
    \State \textbf{Step 2:} Compute the corners of all elements in the sparse mesh
    \State \textbf{Step 3:} Using the corners, compute the areas($A$)  of all elements in the reference dense mesh
    \State \textbf{Step 4:} Compute the thermal flux ($\phi_j = \dfrac{Q_D}{A}$) for each element $j$ in the dense mesh. 
    \For{Each element $i$ in the sparse mesh}
    \State \textbf{Step 5:} Compute the area of overlap($A_c$) with elements in the dense mesh 
    \State \textbf{Step 6:} Interpolated Load $Q_S[i] = \sum_{j=0}^n A_c[i] \times \phi_j $, where n is the number of dense elements a given sparse element intersects.
    \EndFor
\EndProcedure
\end{algorithmic}
\end{algorithm}

\subsubsection{Upscaling Layer}
The upscaling layer (UL) will be a Radial Basis Function (RBF), as this is well-suited for processes such as interpolation of image-like data, as indicated by \cite{Skala2016RBF}. The inputs of this will be the coarse grid temperature profile that is produced from the finite difference model. This produces the final temperature field on the fine grid. Equation \ref{eq: RBF} is the generic form of an RBF. In this paper, a Gaussian basis function is used.
\begin{equation}\label{eq: RBF}
    Y  = \sum_{i=1}^{m} w_i \psi(|| X - X_i||)
\end{equation}
\noindent where:
\begin{description}
\item  $Y$ is the fine mesh temperature profile output
\item $X$ is the coarse grid temperature profile input
% \item $\Theta$ represents the activation function
\item $\psi$ is the basis function 
\item $w$ are the weights
\end{description}

\subsubsection{Loss Functions}
Similar to the process described in \cite{oddiraju2024piml}, the TNN will be trained in a semi-supervised way, constraining the network to achieve the optimal nodalization. Two losses will be implemented, the first is a mean square error loss ($\mathcal{L}_m$), used to match predictions with the data, and the second is a cost loss ($\mathcal{L}_c$), used to encourage the least amount of nodes as possible. The following equation lists the loss functions: 
\clearpage
\begin{equation}\label{eq: total_loss }
    \mathcal{L} = \alpha \mathcal{L}_m + \beta\mathcal{L}_c
\end{equation}
\begin{equation} \label{eq: loss_m}
    \mathcal{L}_m = \frac{1}{D}\sum_{d=1}^{D} w_d\left(\frac{1}{N}  \sum_{i = 1}^{N} ( T_{i,d} - \hat{T}_{i,d})^2
    \right)
\end{equation}

\begin{equation}\label{eq: loss_c}
\begin{split}
 &   \mathcal{L}_c= \frac{1}{N}  \sum_{i = 1}^{N} 10^{4\kappa_i}  \\
 & \kappa_i = \frac{(\sum_{j=1}^J \tau_{i,j}) - \kappa_l }{\kappa_u - \kappa_l} -1   
\end{split}
\end{equation}

\noindent where:
\begin{description}
\item $\alpha, \beta$ are the loss function combination weights 
\item $N$ is the total number of training samples
\item $w_d$ are the assigned weights
\item  $T_{i,d}$ and $\hat{T}_{i,d}$ are the truth and predicted temperatures, corresponding to node $d$ and the $i^{th}$ training sample
\item  $J$ is the total number of body parts on the rover
\item $D$ is the total number of nodes per sample
\item $\tau_{i,j}$ represents the nodalization (number of nodes per dimension  per part) for the $j^{th}$ surface and $i^{th}$ sample
\item $\kappa_i$ is the normalized signed difference between predicted nodalization and the set limit on the total number of nodes per sample
\end{description}

\section{Lunar Rover Case Study} \label{S:casestudy}
\subsection{Model Rover}
\begin{figure}[ht!]
    \centering
    \begin{subfigure}[t]{0.4\textwidth} % <-- top aligned
        \centering
        \includegraphics[width=\linewidth]{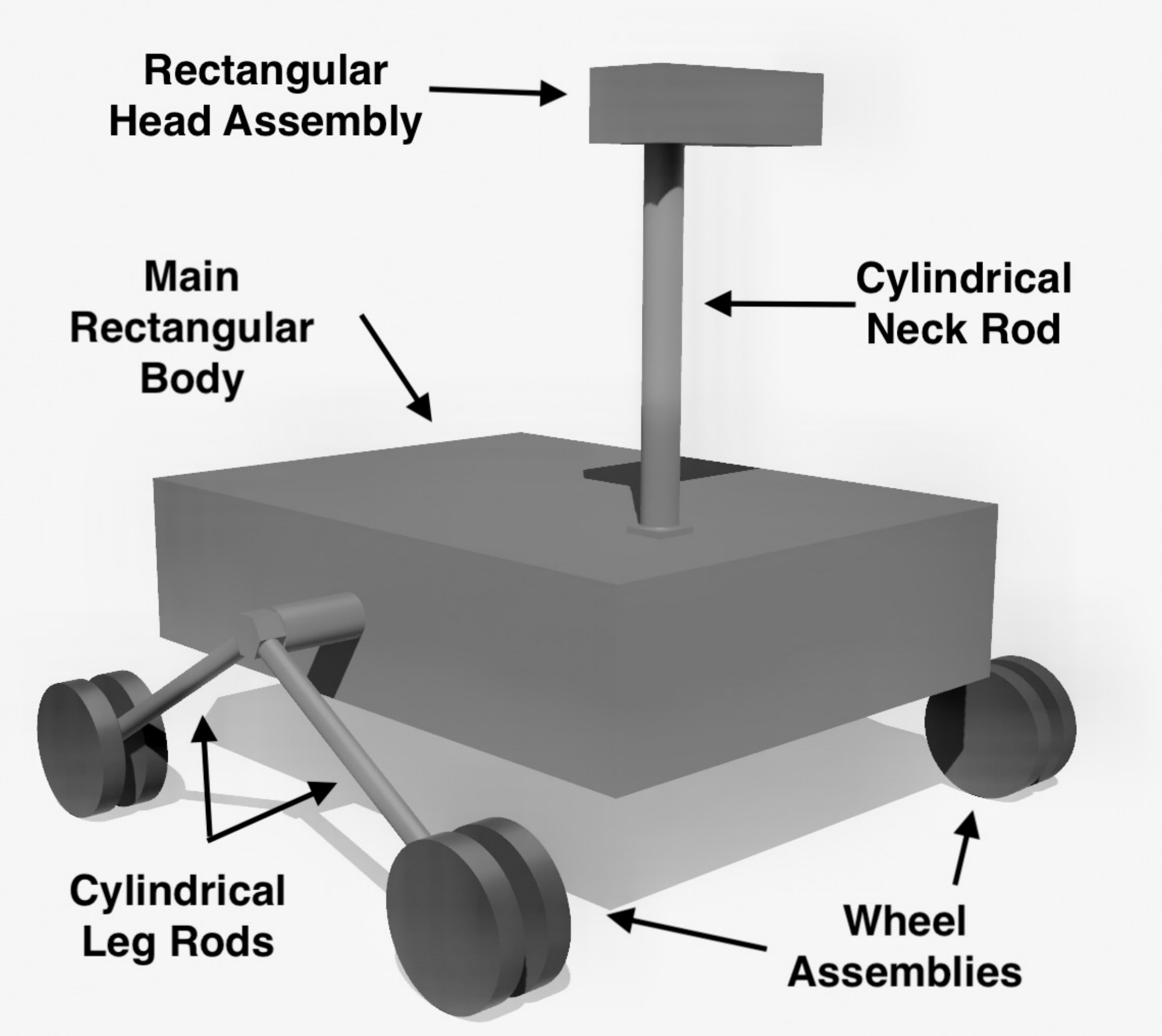}
        \caption{Labeled 3D model of lunar rover}
        \label{fig:rover_cad}
    \end{subfigure}
    \hfill
    \begin{subfigure}[t]{0.45\textwidth} % <-- top aligned
        \centering
        \includegraphics[width=\linewidth]{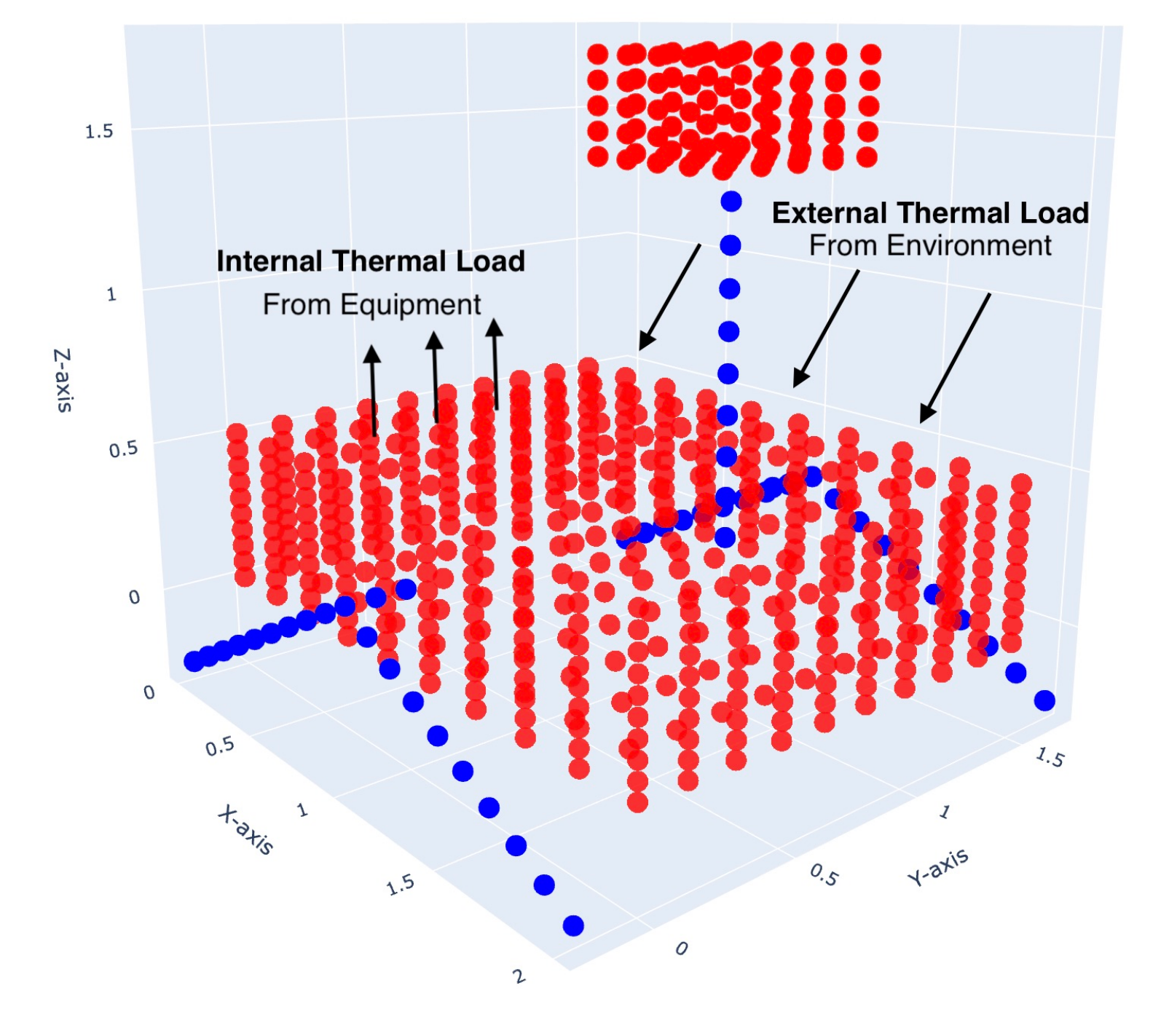}
        \caption{Example nodalization of lunar rover with internal and external loads}
        \label{fig:rover_node}
    \end{subfigure}
    \caption{Simplified lunar rover model used in the case study}
    \label{fig:rover}
\end{figure}

As introduced previously, we are using a simplified model rover, with the 3D rendering shown in Fig. \ref{fig:rover_cad}. The model of the rover consists of a main rectangular body with extruding cylindrical rods that connect to the wheel assembly, and another cylindrical rod that attaches a head assembly filled with equipment. The details of the rover were chosen to provide sufficient complexity to the training dataset, whereas remaining generalized enough to apply to various rover configurations. For this study, thermal analysis of the wheel assembly will be neglected and assumed to maintain a constant temperature. More detailed rovers, with additional extremities and irregular geometries, and increased consideration for internal and external loads, will be studied in future work. Further, such a simplified rover model may be sufficient for onboard planning purposes. Figure \ref{fig:rover_node} shows an example nodalization of the rover, with the red points indicating the rectangular features, and the blue the cylindrical ones. Additionally, the wheels were excluded from this rendering as per the assumption. A summary of the geometry and material properties is found in Table \ref{tab:rover_properties}.

\begin{table}[h!]
\centering
\caption{Rover Geometry and Material Properties}
\begin{tabular}{c c c}
\toprule
\textbf{Parameter} & \textbf{Value} & \textbf{Units} \\
\midrule

\multicolumn{3}{c}{\textit{Geometry}} \\
\midrule
Main Body Dimensions  & $2.0 \times 1.5 \times 0.5$ & m \\
Head Dimensions  & $0.5 \times 0.5 \times 0.25$   & m \\
Neck Length & $1.0$ & m \\
Leg Length  & $1.05$ & m \\
Leg Connector Length & $0.10$  & m \\
\midrule
\multicolumn{3}{c}{\textit{Material Properties}} \\
\midrule
Density ($\rho$) & 2698.79 & kg/m$^3$ \\
Thermal Conductivity ($k$) & 167.304 & W/(m$\cdot$K) \\
Specific Heat ($c_p$) & 895.975 & J/(kg$\cdot$K) \\
Emissivity ($\epsilon$) & 0.82 & --\\
\bottomrule
\end{tabular}
\label{tab:rover_properties}
\end{table}

\subsection{Finite Difference Formulation}
 Equations \eqref{eq:GoverningEq} and \eqref{eq:ForwardEuler} show a 3D lumped thermal mass heat transfer formulation with terms of the individual nodal heat transfer ($q_{i,j}$) \cite{jaluria2002heat}. Equation \eqref{eq:GoverningEq} is a version of the Fourier-Biot equation, which serves as the governing equation for this formulation. Equation \eqref{eq:ForwardEuler} is a forward Euler discretization of Eq. \eqref {eq:GoverningEq}. The nodalization depends on the spacings between nodes and will be decided by the PIML model. An explicit formulation was employed. Both the internal and external loads are included within the $q_{i,j,k}$ term. 
% \clearpage
\begin{equation} \label{eq:GoverningEq}
\frac{\partial T}{\partial t}
=
\alpha \left(
\frac{\partial^2 T}{\partial x^2}
+
\frac{\partial^2 T}{\partial y^2}
+
\frac{\partial^2 T}{\partial z^2}
\right)
+
\frac{q_{i,j,k}}{\rho c}
\end{equation}

\begin{equation} \label{eq:ForwardEuler}
\begin{split}
\frac{T_{i,j,k}^{n+1} - T_{i,j,k}^{n}}{\Delta t} 
= \alpha \Bigg(
& \frac{T_{i+1,j,k}^{n} - 2T_{i,j,k}^{n} + T_{i-1,j,k}^{n}}{\Delta x^2} \\
& + \frac{T_{i,j+1,k}^{n} - 2T_{i,j,k}^{n} + T_{i,j-1,k}^{n}}{\Delta y^2} \\
& + \frac{T_{i,j,k+1}^{n} - 2T_{i,j,k}^{n} + T_{i,j,k-1}^{n}}{\Delta z^2}
\Bigg) \\ &
+  \frac{q_{i,j,k}}{\rho c}
\end{split}
\end{equation}

\noindent where:
\begin{description}
\item $\alpha$ is the thermal conductivity
\item $c$ is the specific heat
\item $n$ is the current time iteration
\item $\rho$ is the material density
\item $T_{i,j,k}$ is the temperature at node $(i,j,k)$
\item $\Delta t$ is the time step
\item $q_{i,j,k}$ is the net heat transfer rate of internal and external loads
\item $\Delta x, \Delta y, \Delta z$ are the spacing between nodes
\end{description}

\subsection{Boundary and Initial Conditions}
For this paper, the modeling problem is directed towards predicting the temperature profile of all surfaces given the initial temperature distribution and thermal loads. In this paper, we will start at a uniform temperature distribution set to equal the ambient surface temperature of the moon, using data found in \cite{hamill2021lunar}. For boundary conditions, we will assume that only radiation heat transfer is occurring and the ambient temperature remains constant over the simulation time. Stefan–Boltzmann law will be used to model this.

\begin{equation} \label{eq:SBradiation}
q_{rad_ {i,j,k} }= \varepsilon \sigma A_{i,j,k} \left( T_{i,j,k}^4 - T_{\text{amb}}^4 \right)
\end{equation}

\subsection{Thermal Loads}
Both external and internal loads will be considered. External loads will be in the form of a solar flux and will be based on the orientation of each surface to the sun. The sun will be modeled as a single point in space, defined by elevation and azimuth angles, $\phi$ and $\theta$, respectively. Each node will calculate the fraction of the radiation that it will receive to ensure shadowing effects from other rover components are as accurate as possible. The values of these solar fluxes will be estimated using data from the Apollo mission and other lunar data found in \cite{hamill2021lunar}. A higher accuracy model taking into account orbit parameters is a possible subject of future research.

Internal loads will be represented by known heat transfer rates, indicative of phenomena such as joule heating or internal heating sources. For this paper, we will implement 2 heat sources in the main body and 1 in the head assembly, and vary the position and strength of each. This represents various equipment needed for safe operation, as well as scientific equipment payloads. A summary of all the inputs and outputs is shown in Table \ref{tab:in_out}. 

\begin{table*}[h!]
\centering
\caption{Summary of Inputs and Outputs}
\begin{tabular}{c c c c}
\toprule
\textbf{Input/Output} &\textbf{Parameter} & \textbf{Bounds} & \textbf{Dimensionality}\\
\midrule
\multirow{5}{*}{Inputs} & Incoming Solar Flux  &  1315 -- 1421 $\frac{W}{m^2}$   &1  \\
& Elevation Angle  &0 -- 90$^\circ$  &1     \\
& Azimuth Angle & 0 -- 360$^\circ$    & 1      \\
& Heat Source Power ($\times$ 3) & 100 -- 2000 $W$ & 1  \\
& Heat Source Location ($\times$ 3) & Body \& Head Bounds & 3 \\
       \cmidrule{1-4}
\multirow{1}{*}{Outputs} & Normalized Fine Mesh Temperatures  & 0 -- 1   & 1  \\

\bottomrule
\end{tabular}
\label{tab:in_out}
\end{table*}

\section{Results and Discussion} \label{S:results}

This section presents the results of the proposed physics-informed machine learning (PIML) framework through comparisons with data-driven and physics-based baselines, evaluating both predictive accuracy and computational performance.

\subsection{Baselines}

\begin{figure}[ht!]
    \centering
    \begin{subfigure}[t]{0.49\textwidth} % <-- top aligned
        \centering
        \includegraphics[width=\linewidth]{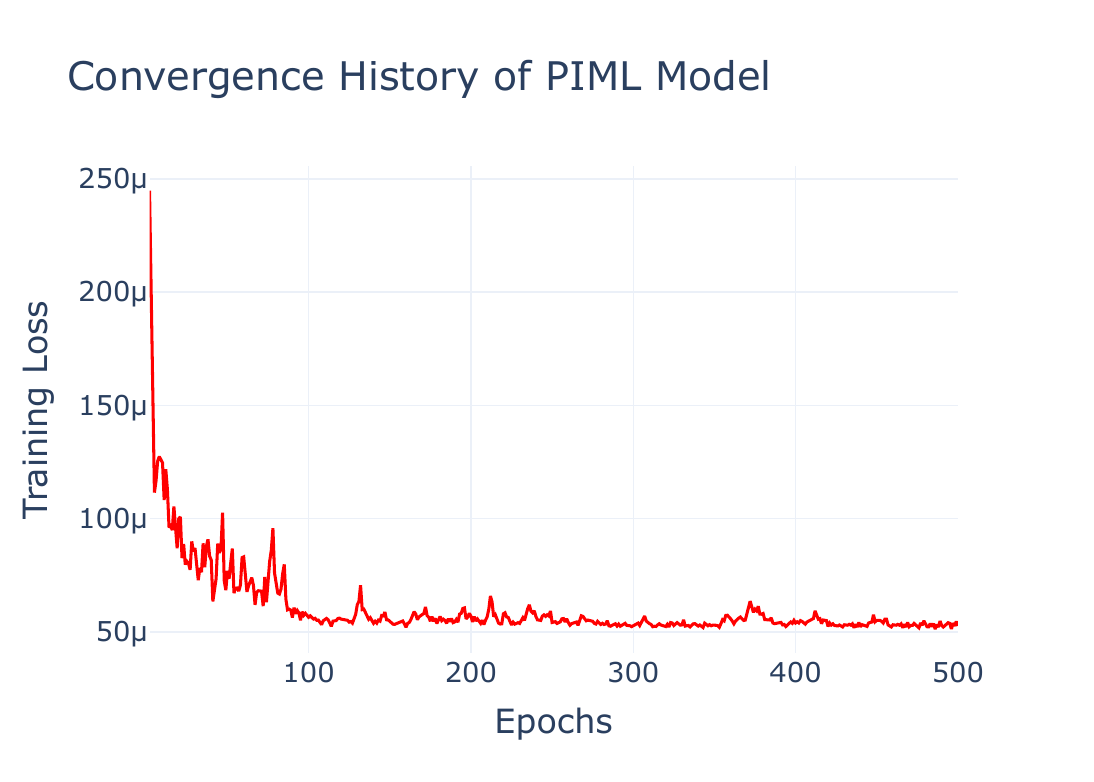}
        \caption{PIML training convergence history}
        \label{fig:piml_conv}
    \end{subfigure}
    \hfill
    \begin{subfigure}[t]{0.49\textwidth} % <-- top aligned
        \centering
        \includegraphics[width=\linewidth]{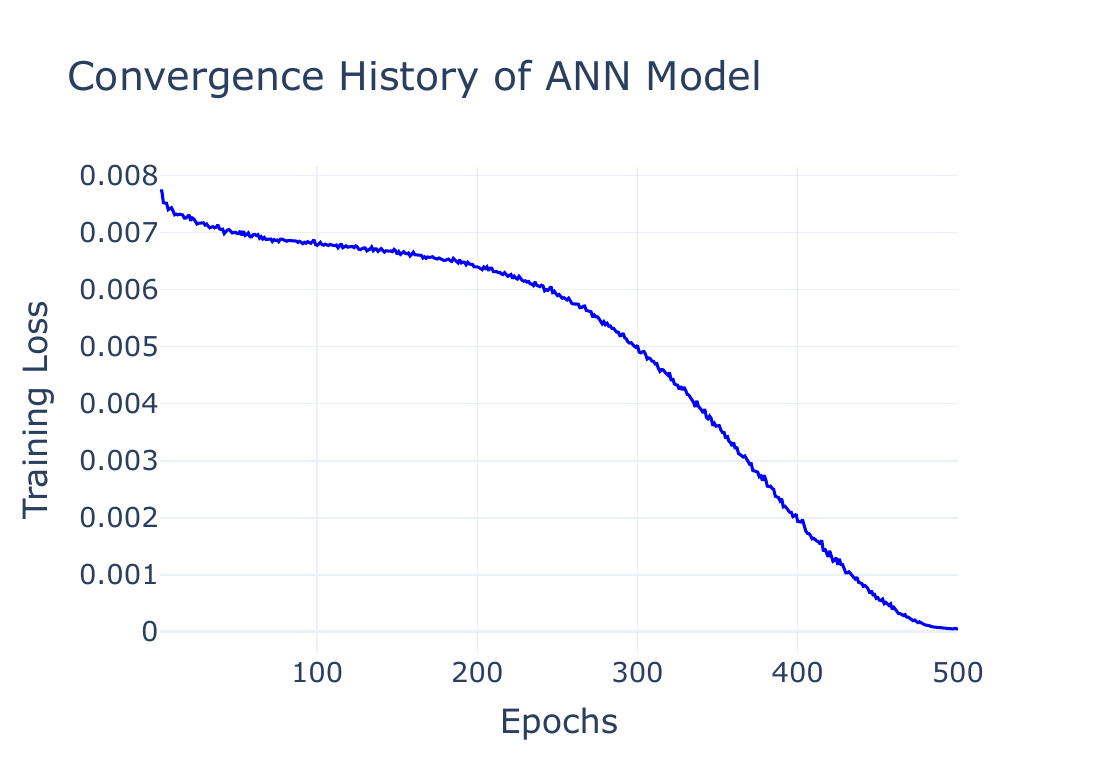}
        \caption{ANN training convergence history}
        \label{fig:ann_conv}
    \end{subfigure}
    \caption{Training convergence histories for PIML and ANN models }
    \label{fig:converge}
\end{figure}

To evaluate performance, the PIML model is compared against both data-driven and physics-based baselines. The data-driven model consists of an artificial neural network (ANN) trained using a mean squared error (MSE) loss function. Note that the ANN here predicts the entire temperature field, at the same grid resolution as the high-fidelity FD model (for ease of comparisons and loss computations). The ANN is intentionally kept simple to serve as a lower bound reference. More advanced architectures, such as generative adversarial networks (GANs) or encoder–decoder models, may be better suited for high-dimensional field prediction but are not considered here. The configurations of the ANN and the PIML transfer neural network (TNN) are summarized in Table \ref{tab:NN_Param}. Both the PIML and ANN models are allowed to train till 500 epochs, with the training convergence histories shown in Fig. \ref{fig:converge}. We observe that PIML demonstrates some early oscillations (likely artifacts of the backpropagation through FD) in training but overall faster convergence, while ANN shows a smoother but overall slower convergence. 

The physics-based baselines are derived from a uniform-mesh implementation of the internal thermal simulator and differ only in spatial resolution. The low-fidelity (LF) model provides a coarse approximation, while the high-fidelity (HF) model is treated as the ground truth solution. The nodal distributions for each rover component are listed in Table \ref{tab:LF_HF}, selected to ensure consistent spatial discretization across all components.

Model inputs include environmental thermal loads, rover-to-sun orientation, and internal heat source locations and magnitudes, in part motivated by \cite{hamill2021lunar,quattrocchi2022curiosity}. The ANN directly predicts the full temperature field, whereas the PIML model predicts nodalization parameters, which are then used as inputs to the thermal simulator.  

%Configs
\begin{table}[h!]
\centering
\caption{Neural Network Configurations for ANN and PIML Models}
\begin{tabular}{c c c}
\toprule
\textbf{Parameter} & \textbf{ANN Value} & \textbf{PIML Value} \\
\midrule

Number of Hidden Layers & 5 & 5 \\
Number of Nodes Per Layer & 1024 & 1024 \\
Learning Rate & $10^{-3}$ & $10^{-3}$ \\
Number of Training Samples & 1000 & 1000 \\
Number of Testing Samples & 200 & 200 \\
Sampling Approach & LHS (max-min) & LHS (max-min)\\
Number of Epochs & 500 & 500 \\
Activation Function (Hidden Layers) & Leaky ReLU & Leaky ReLU \\
Number of Inputs & 15 & 15 \\
Number of Outputs & 14768 & 52 \\

\bottomrule
\end{tabular}
\label{tab:NN_Param}
\end{table}

\begin{table}[h!]
\centering
\caption{Nodalization for High-Fidelity (FD) and Low-Fidelity (FD) Physics Models}
\begin{tabular}{c c c c}
\toprule
\textbf{Component} & \textbf{Direction} & \textbf{High-Fidelity} & \textbf{Low-Fidelity} \\
\midrule

\multirow{3}{*}{Main Body}
& x & 41 & 9 \\
& y & 31 & 7 \\
& z & 11 & 3 \\
\cmidrule{1-4}

\multirow{3}{*}{Head Assembly}
& x & 11 & 5 \\
& y & 11 & 5 \\
& z & 6 & 3 \\
\cmidrule{1-4}

Neck & -- & 11 & 3 \\
Leg (Each) & -- & 12 & 3 \\
Leg Connection (Each) & -- & 3 & 3 \\

\bottomrule
\end{tabular}
\label{tab:LF_HF}
\end{table}

\subsection{Performance and Interpretability}
\begin{figure}[ht!]
    \centering
    \begin{subfigure}[t]{0.44\textwidth} % <-- top aligned
        \centering
        \includegraphics[width=\linewidth]{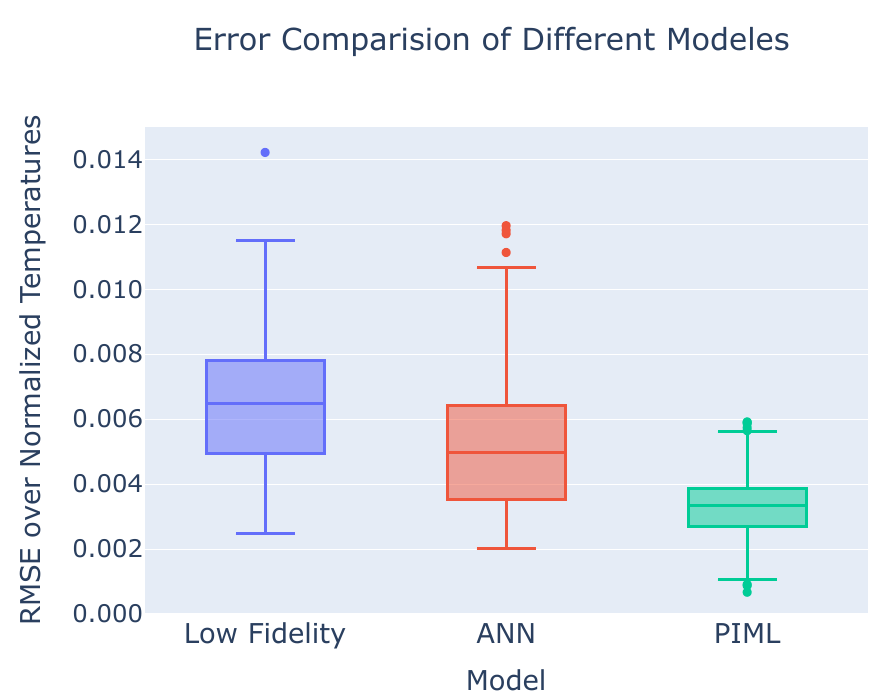}
        \caption{Box plot comparison of prediction errors to the high-fidelity FD model}
        \label{fig:box}
    \end{subfigure}
    \hfill
    \begin{subfigure}[t]{0.44\textwidth} % <-- top aligned
        \centering
        \includegraphics[width=\linewidth]{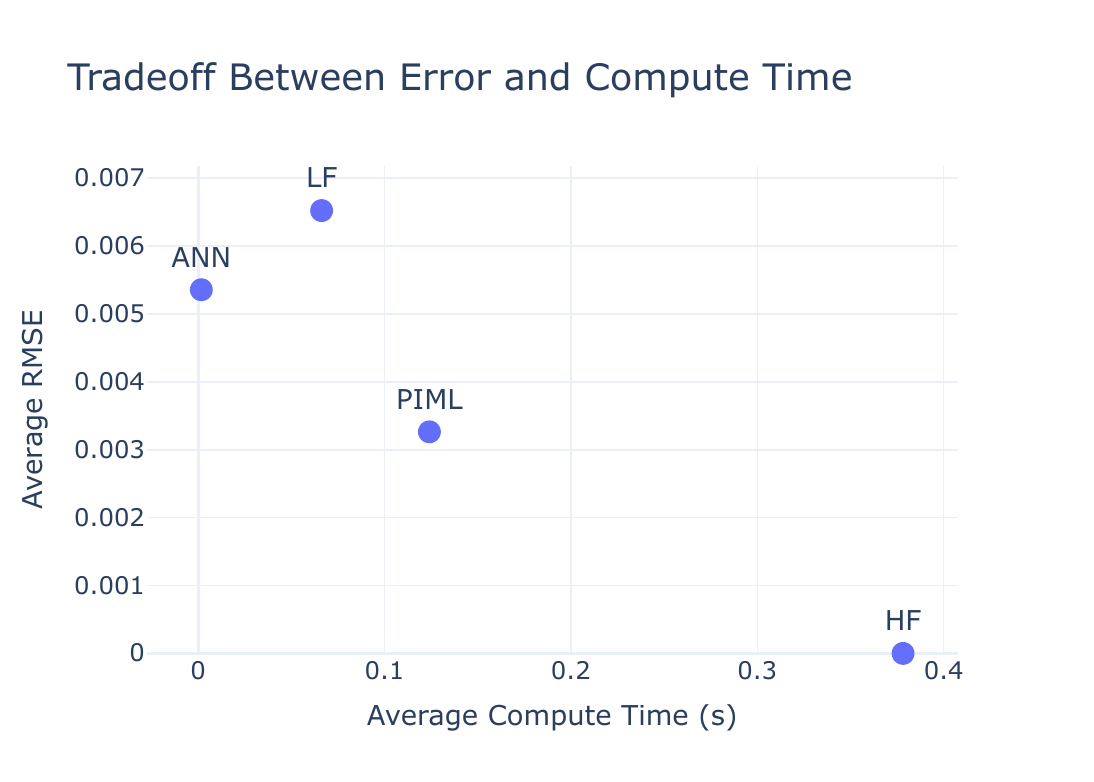}
        \caption{Tradeoff between prediction errors and average compute time}
        \label{fig:tradeoff}
    \end{subfigure}
    \caption{Comparison of model performances}
    \label{fig:comparisions}
\end{figure}

Figure \ref{fig:box} presents the root mean square error (RMSE) comparison for the LF, ANN, and PIML models over the 200 test samples. The PIML framework achieves the lowest overall error, demonstrating clear improvement over both the LF and ANN approaches. The ANN provides only marginal improvement over the LF model, indicating limited generalization under the available training data. The PIML reduces the mean RMSE by approximately 50\% relative to the LF model and by 39\% relative to the ANN. The mean error values are also reported in Table \ref{tab:errors}. In addition, note from the figures that the PIML model demonstrates significantly lower variance, thus better robustness, compared to the LF and ANN models. Figure \ref{fig:tradeoff} shows a trade-off analysis of the four models in terms of average RMSE values (w.r.t. the HF model, which is considered the ground truth here) and average compute times (which is discussed further later in this section). This figure clearly brings out how PIML offers an attractive trade-off in terms of these two metrics.

% \begin{figure}[ht!]
%     \centering
%     \begin{subfigure}[t]{0.4\linewidth}
%         \centering
%         \includegraphics[width=\linewidth]{FIGURES/BOX.pdf}
%         \caption{Box plot comparison of prediction errors to the high-fidelity FD model}
%         \label{fig:box}
%     \end{subfigure}
%     \hfill
%     \begin{subfigure}[t]{0.45\linewidth}
%         \centering
%         \includegraphics[width=\linewidth]{FIGURES/Tradeoff.pdf}
%         \caption{Tradeoff between prediction errors and average compute time}
%         \label{fig:tradeoff}
%     \end{subfigure}
%     \caption{Comparison of model performances}
%     \label{fig:comparisions}
% \end{figure}

\begin{table}[ht!]
\centering
\caption{Average Root Mean Square Error of Each Model}
\begin{tabular}{c c}
\toprule
\textbf{Model} & \textbf{Average RMSE} \\
\midrule

LF & 0.006523 \\
ANN & 0.005357 \\
PIML & \textbf{0.003264} \\

\bottomrule
\end{tabular}
\label{tab:errors}
\end{table}

\begin{figure}[h!]
    \centering

    \begin{subfigure}[b]{0.49\linewidth}
        \centering
        \includegraphics[width=\linewidth]{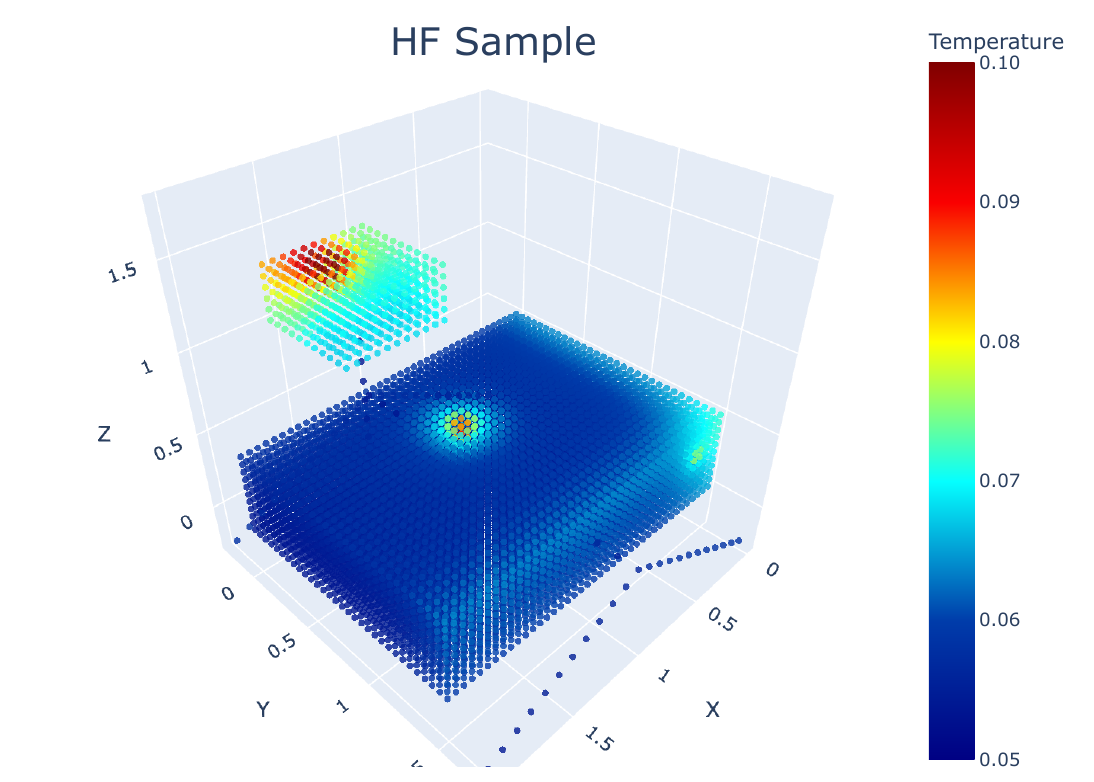}
        \caption{High-Fidelity Model}
    \end{subfigure}
    \hfill
    \begin{subfigure}[b]{0.49\linewidth}
        \centering
        \includegraphics[width=\linewidth]{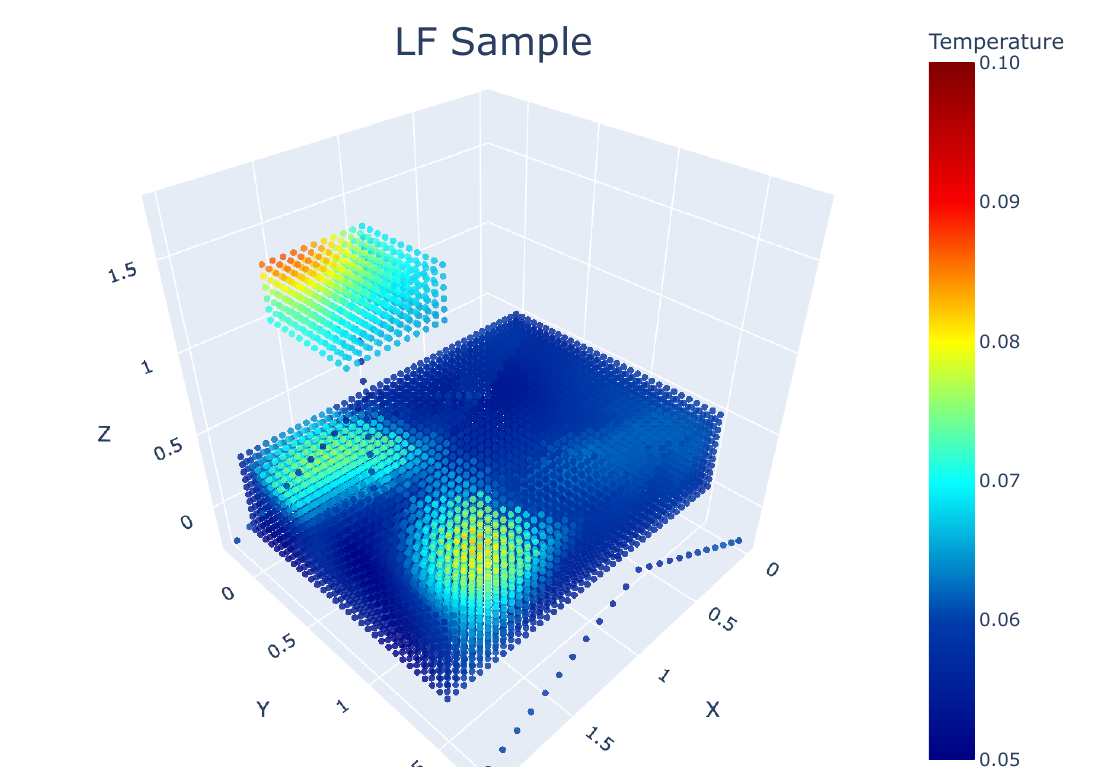}
        \caption{Low-Fidelity Model}
    \end{subfigure}

    \vspace{0.3cm}

    \begin{subfigure}[b]{0.49\linewidth}
        \centering
        \includegraphics[width=\linewidth]{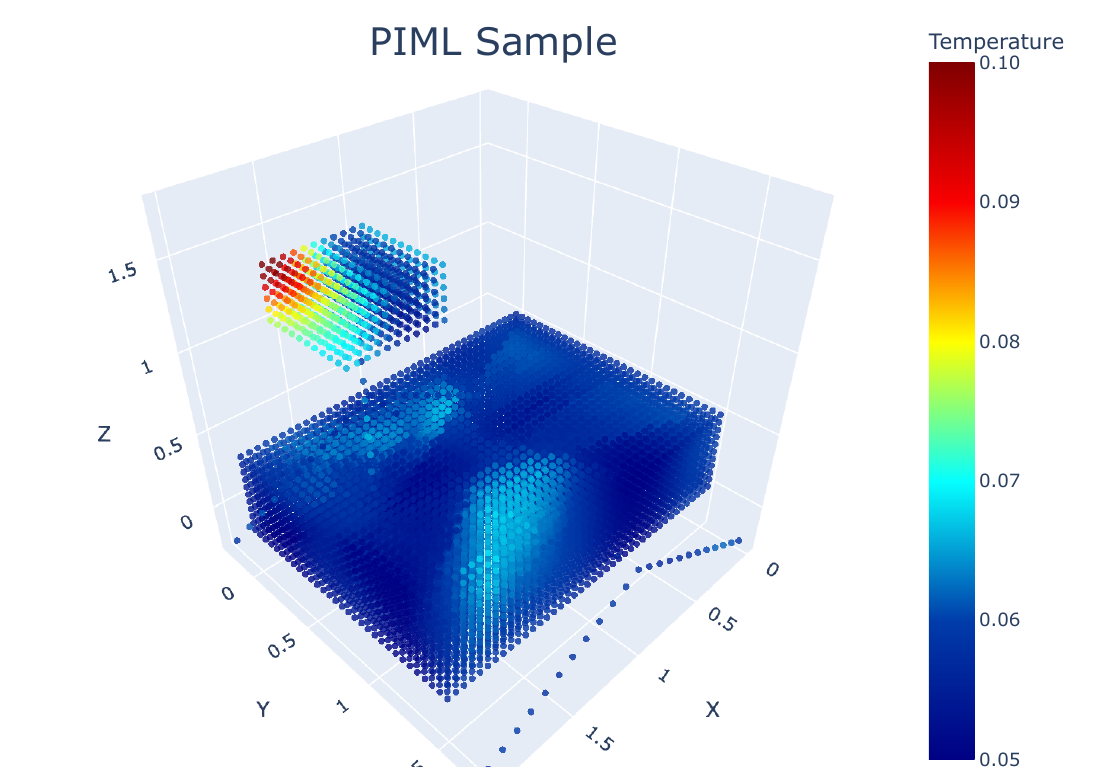}
        \caption{PIML}
    \end{subfigure}
    \hfill
    \begin{subfigure}[b]{0.49\linewidth}
        \centering
        \includegraphics[width=\linewidth]{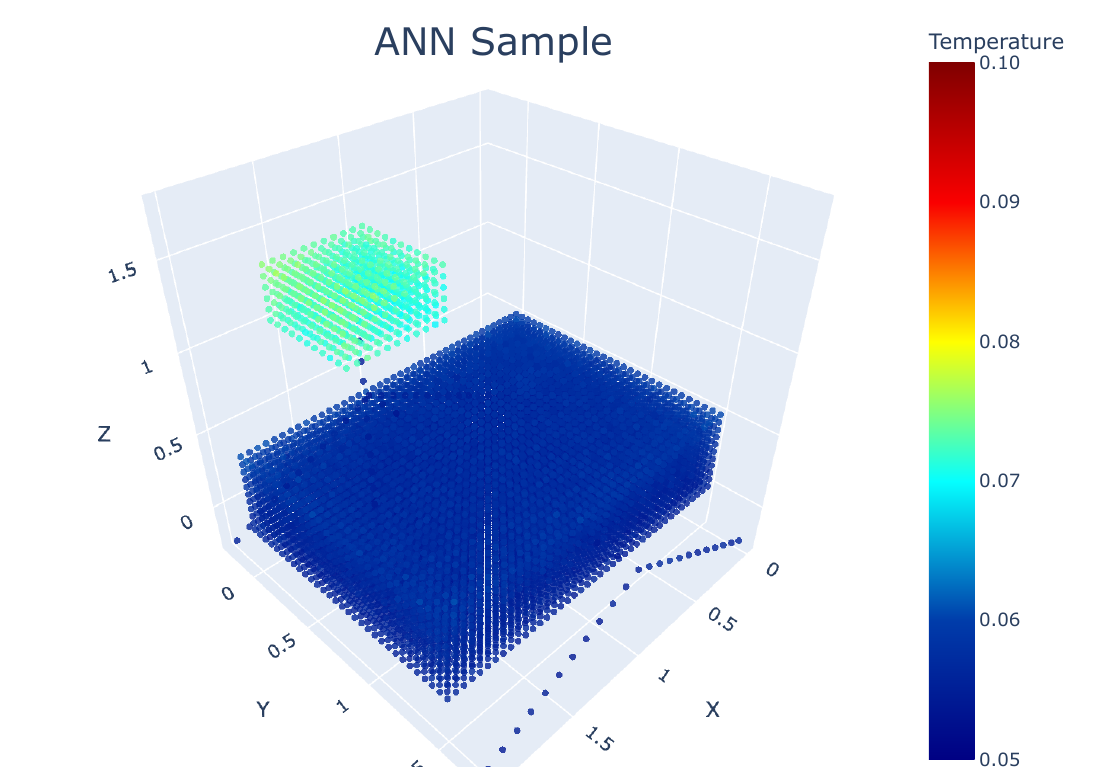}
        \caption{ANN Model}
    \end{subfigure}

    \caption{Final temperature distributions of each Model for representative test case 1 }
    \label{fig:Ex2}
\end{figure}

\begin{figure}[h!]
    \centering

    \begin{subfigure}[b]{0.49\linewidth}
        \centering
        \includegraphics[width=\linewidth]{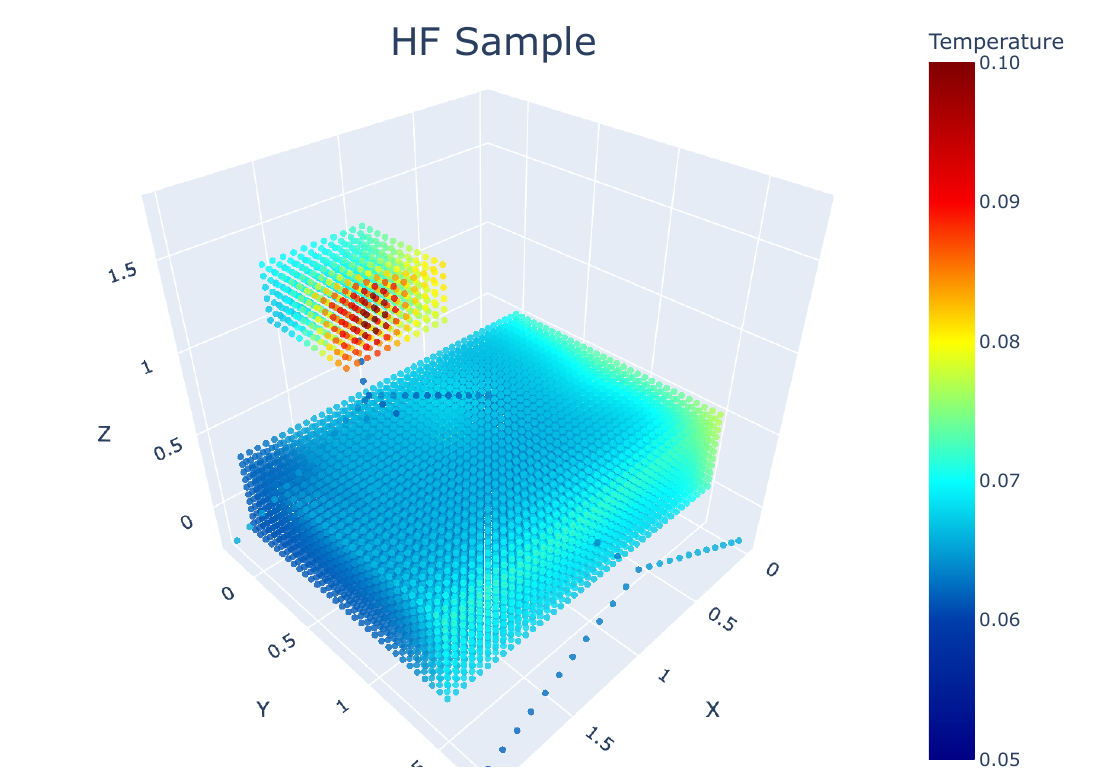}
        \caption{High-Fidelity Model}
    \end{subfigure}
    \hfill
    \begin{subfigure}[b]{0.49\linewidth}
        \centering
        \includegraphics[width=\linewidth]{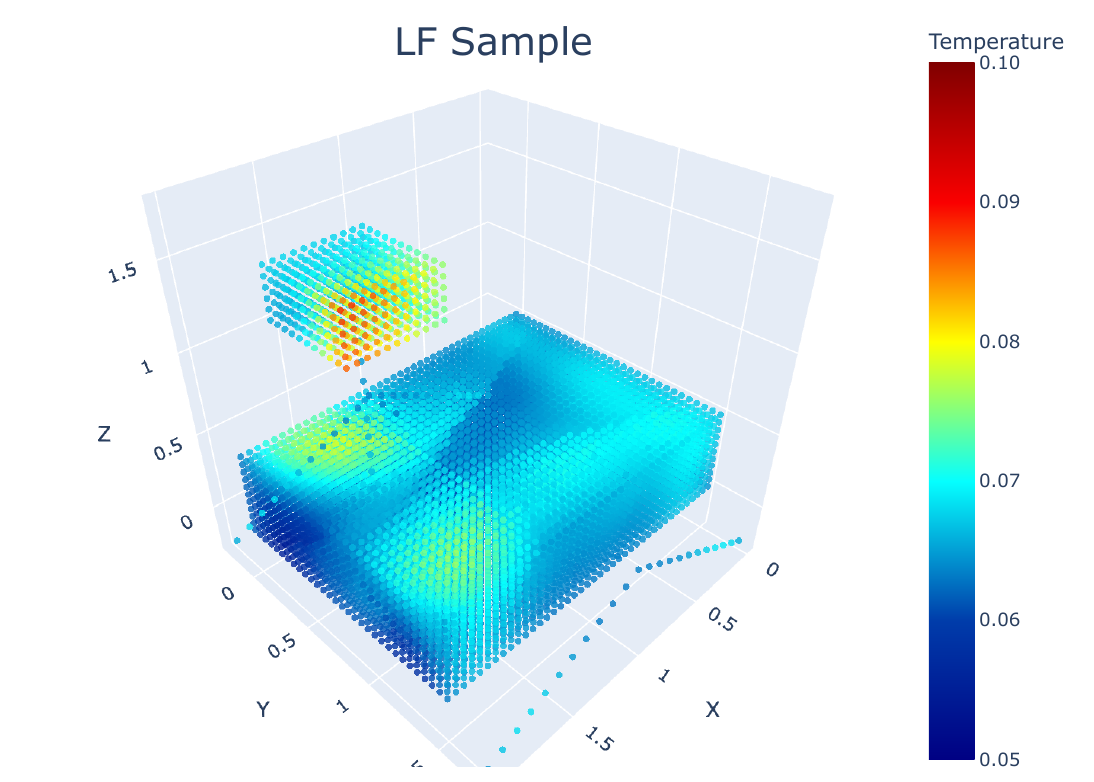}
        \caption{Low-Fidelity Model}
    \end{subfigure}

    \vspace{0.3cm}

    \begin{subfigure}[b]{0.49\linewidth}
        \centering
        \includegraphics[width=\linewidth]{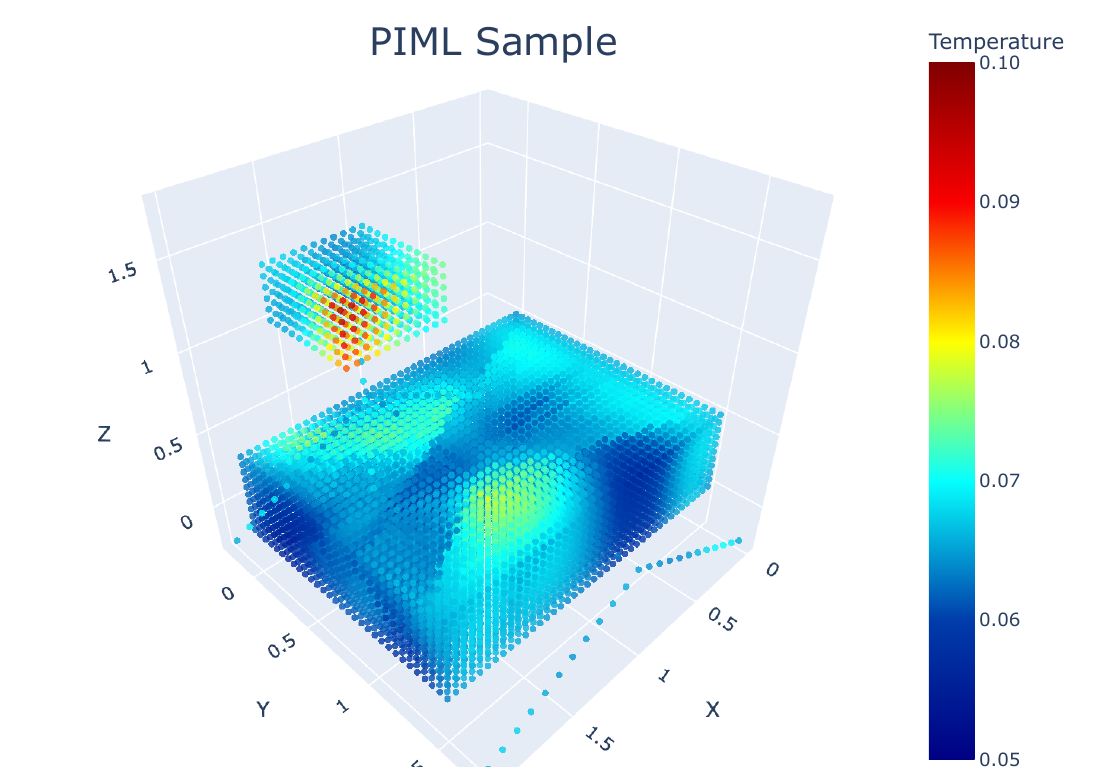}
        \caption{PIML}
    \end{subfigure}
    \hfill
    \begin{subfigure}[b]{0.49\linewidth}
        \centering
        \includegraphics[width=\linewidth]{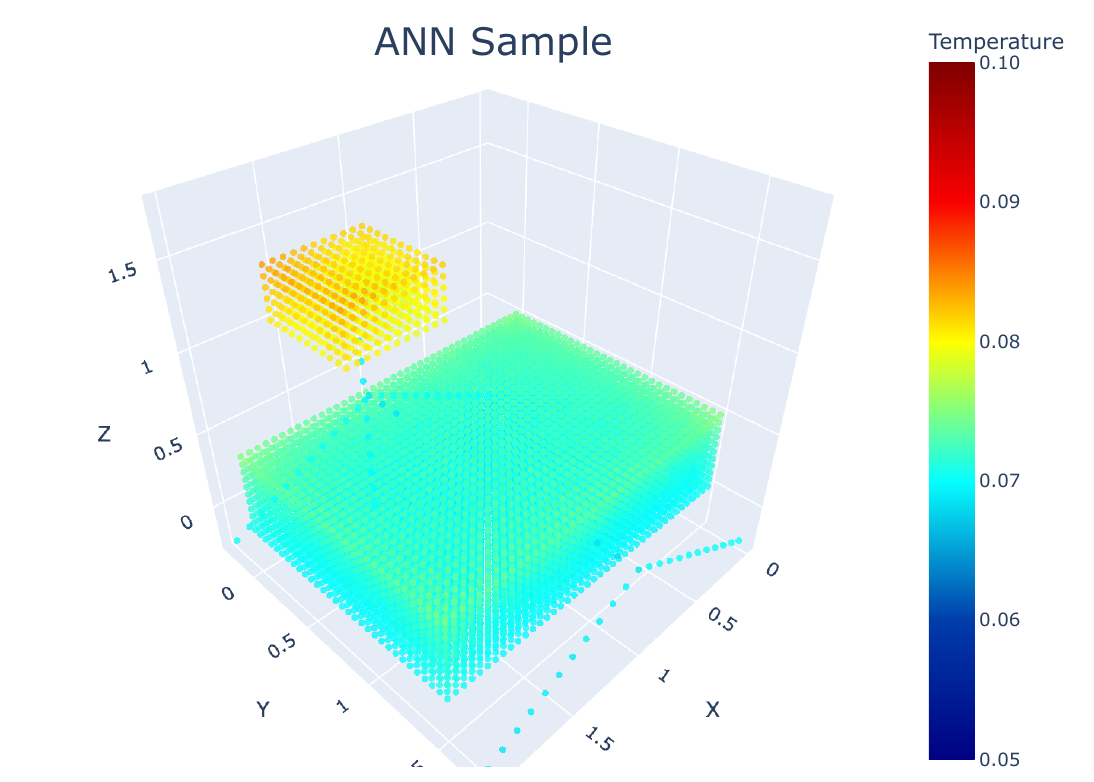}
        \caption{ANN Model}
    \end{subfigure}

    \caption{Final temperature distributions of each Model for representative test case 2 }
    \label{fig:Ex1}
\end{figure}

We now choose two representative test cases to assess the temperature predictions made by each method, and associated implications, e.g., in terms of grid density. These cases were randomly sampled from the 200 testing samples. The predicted temperature contours are shown in Figs.~\ref{fig:Ex2} and Fig.~\ref{fig:Ex1} respectively for the two cases. The mean errors for these two cases are reported in Table \ref{tab:case_errors}. In both cases, the PIML model is observed to more accurately capture the thermal field distribution than the ANN and LF models, although localized discrepancies remain. These discrepancies are most pronounced near heat sources, where reduced nodalization limits spatial resolution. Despite this, both the PIML and LF models produce physically consistent thermal distributions, whereas the ANN exhibits less realistic spatial behavior. This limitation in the ANN predictions is likely attributable to the relatively small training dataset in relation to the cardinality of the input space and especially the output space (field predictions).

\begin{table}[ht!]
\centering
\caption{Average Error of Each Model for Two Test Cases}
\begin{tabular}{c c c}
\toprule
\textbf{Model} & \textbf{Average RMSE -- Case 1} & \textbf{Average RMSE -- Case 2} \\
\midrule

LF & 0.005904 & 0.006285 \\
ANN & 0.004260 & 0.004558 \\
PIML & \textbf{0.003668} & \textbf{0.003204} \\

\bottomrule
\end{tabular}
\label{tab:case_errors}
\end{table}

\begin{figure}[ht!]
    \centering

    \begin{subfigure}[t]{0.8\linewidth}
        \centering
        \includegraphics[width=\linewidth]{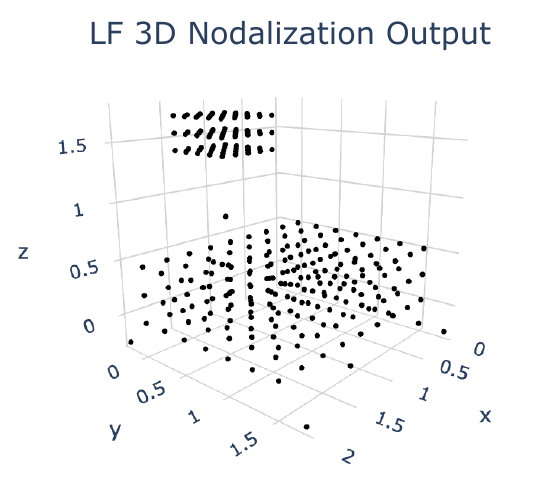}
        \caption{Mesh of LF model}
        \label{fig:comp_lf}
    \end{subfigure}

    \vspace{0.3cm}

    \begin{subfigure}[t]{0.8\linewidth}
        \centering
        \includegraphics[width=\linewidth]{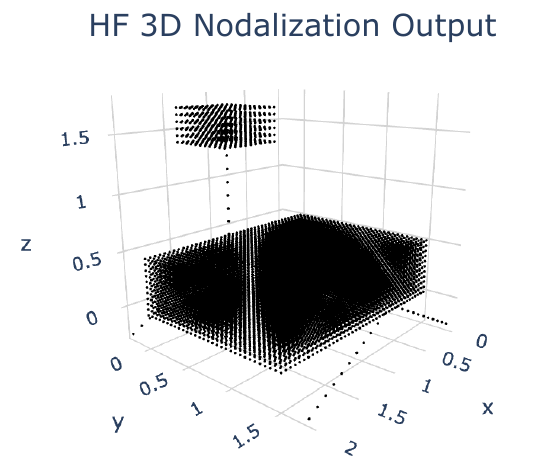}
        \caption{Mesh of HF model}
        \label{fig:comp_hf}
    \end{subfigure}

    \vspace{0.3cm}

    \begin{subfigure}[t]{0.8\linewidth}
        \centering
        \includegraphics[width=\linewidth]{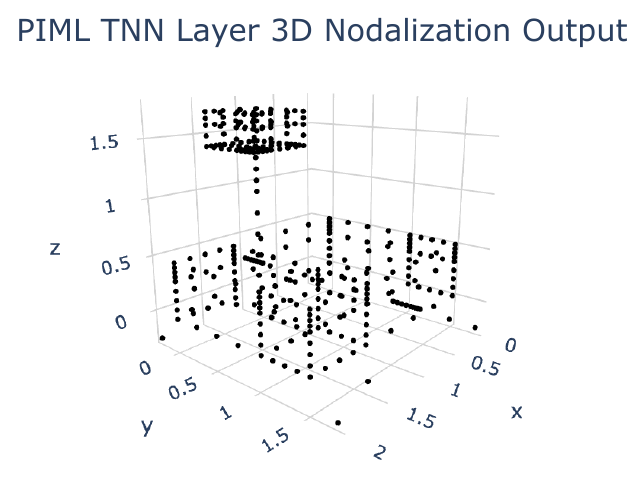}
        \caption{Mesh created from TNN outputs}
        \label{fig:comp_tnn}
    \end{subfigure}

    \caption{Comparison of meshes created from LF, HF, and TNN models}
    \label{fig:comp_out}
\end{figure}

The spatial distribution of these discrepancies suggests that model accuracy is closely tied to how resolution is allocated across the domain. To investigate this, for the example case 2, the fixed mesh used by the LF and HF models, and the mesh adapted by the TNN in the PIML are shown in Fig. \ref{fig:comp_out}. With the TNN in PIML, the resulting mesh is non-uniform, showing higher node density concentrated in smaller geometric regions, such as the head, where steeper temperature gradients are expected. In contrast, larger regions exhibit smoother behavior and allow for coarser discretization in other parts of the rover. Some variation in nodal density is also observed in regions subjected to higher heat flux, suggesting that the TNN may be adapting the discretization in response to both geometric features and loading conditions. This behavior is consistent with physical expectations and suggests that both geometric features and loading conditions influence the learned nodalization.

%temp ex1
%temp ex2

%node compare

\subsection{Compute Cost Analysis}
\begin{figure}[ht!]
    \centering
    \includegraphics[width=0.8\linewidth]{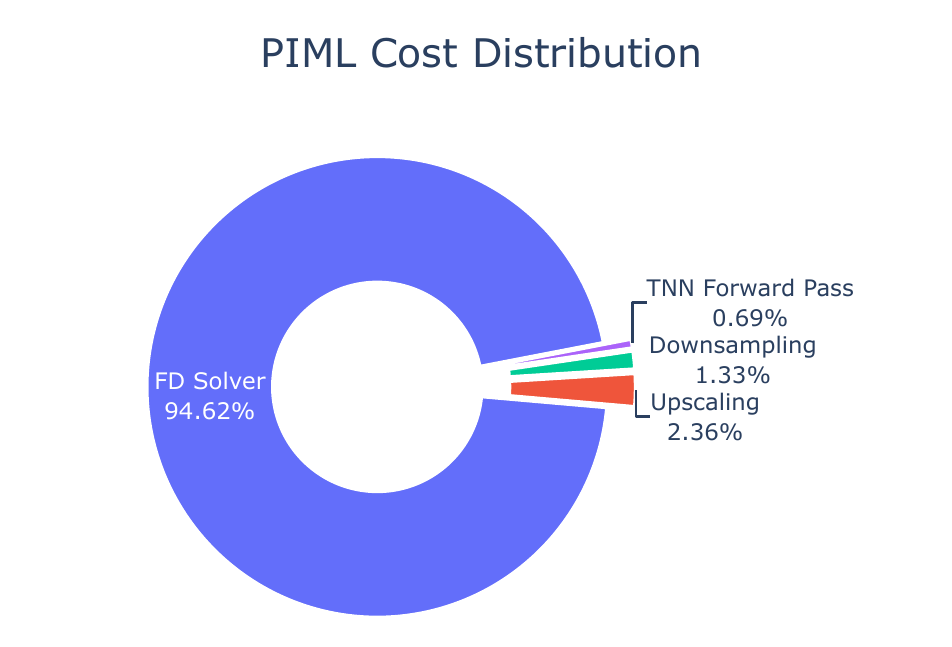}
    \caption{Average runtime cost breakdown for PIML architecture}
    \label{fig:cost_break}
\end{figure}

Computational performance is summarized in Table~\ref{tab:times}. All runtimes were measured on a single workstation equipped with an Apple M4 Max processor (14-core CPU, 30-core GPU, 36 GB unified memory). As expected, the ANN exhibits the lowest runtime due to the absence of any physics-based computations. The PIML framework provides a balance between accuracy and efficiency, with a runtime significantly lower than the HF model and closer to that of the LF model, while maintaining improved predictive accuracy. Specifically, PIML achieves an approximately 3$\times$ speedup relative to the HF model, while incurring only a 1.9$\times$ increase in runtime compared to the LF model. The latter is attributed to the following: the costs of the transfer network, downsampling and upsampling components, as well as the fact that the PIML can have a mesh with grid density anywhere within the range bounded by the LF and HF models' fixed meshes, and thus usually involves slightly higher mesh density on average compared to the standalone LF model. Note here that upscaling is a notable portion of the cost, which could be eliminated during deployment, since the purely LF model does not include that upscaling. Although the ANN is approximately 70$\times$ faster than PIML, this speed comes at the cost of reduced accuracy and diminished physical consistency.
A breakdown of the computational cost within the PIML architecture is shown in Fig.~\ref{fig:cost_break}, where the embedded physics model accounts for the majority of the runtime. Overall, these results demonstrate a favorable accuracy--cost tradeoff, with substantial error reduction relative to the LF model achieved at a modest computational overhead, while remaining significantly faster than the HF baseline.

%cost
\begin{table}[h!]
\centering
\caption{Run Time of Each Model}
\begin{tabular}{c c c}
\toprule
\textbf{Model} & \textbf{Mean Run Time (s)} & \textbf{Standard Deviation (s)} \\
\midrule

HF & 0.3782 & 0.01475 \\
LF & 0.06626 & 0.003416 \\
PIML & 0.1241 & 0.005884 \\
ANN & \textbf{0.001740} & 0.0001324 \\

\bottomrule
\end{tabular}
\label{tab:times}
\end{table}

%Cost breakdown

\section{Concluding Remarks} \label{S:conclusion}

In this work, a physics-informed machine learning (PIML) framework was developed for efficient thermal (temperature profile) prediction of a lunar rover, with a focus on balancing predictive accuracy and computational cost. The proposed architecture integrates a transfer neural network (TNN) for adaptive nodalization and an upscaling layer with a differentiable physics-based simulator implemented using an explicit finite difference scheme in the JAX library. This hybrid formulation enables efficient training while preserving physical consistency and resolution in the predicted temperature fields. Training convergence demonstrated the effectiveness of the end-to-end auto-differentiable implementation of the PIML. Tested on unseen samples, the results demonstrated that the PIML framework significantly improves predictive accuracy relative to a low-fidelity (LF) model, reducing mean RMSE by approximately 50\%, while maintaining physically consistent thermal distributions and only slightly increasing the compute cost. Compared to a purely data-driven artificial neural network (ANN), the PIML approach produces more realistic spatial temperature fields, particularly in regions with strong thermal gradients. In terms of computational performance, the PIML achieves an approximate 3$\times$ speedup relative to the high-fidelity (HF) model. These results highlight the effectiveness of the proposed framework in achieving a favorable accuracy--cost tradeoff. This is an initial implementation demonstrating a proof of concept of this approach to PIML for lunar rover thermal analysis, with scope for significantly greater compute efficiency gains in the future -- supported, for example, by more aggressively bounding the maximum average grid density (tunable by the transfer network), and tuning the time stepping as well using the ML components.  

% Overall, the proposed PIML approach provides a practical and scalable pathway for integrating physics-based modeling with machine learning, enabling rapid and reliable thermal analysis for complex space systems during both design and ground and onboard operations. 
This PIML-based onboard simulation is expected to particularly enable autonomous thermal-aware planning in rovers by reducing computational requirements while maintaining physical accuracy; testing of this in a simulated operations setting is an important direction of future work. In addition, the current implementation is limited to a relatively simple rover geometry, internal structure, and thermal source and loading model; more sophisticated implementations of these factors are needed in the future to support optimization-based active cooling/heating decisions, autonomy decisions, or even design decisions for lunar rovers, further informed by mission-specific environmental parameters. 

% Future work will extend the framework to more complex rover geometries and configurations to further assess generalization capabilities. Additionally, incorporating more realistic boundary conditions, including orbital effects, will be investigated to improve model fidelity and applicability to mission-relevant scenarios.

\bibliographystyle{IEEEtran}
\bibliography{Aviation}

\end{document}